\documentclass[runningheads]{llncs}

\usepackage{amsmath,graphicx}
\usepackage{cite}
\usepackage{acro}
\usepackage{tabularx}
\usepackage{subcaption}
\usepackage{booktabs}
\graphicspath{{Figure/}}
\usepackage{color}
\usepackage{multirow}
\usepackage{threeparttable}

\newcolumntype{Y}{>{\centering\arraybackslash}X}
\newcommand{\ie}{\emph{i.e.},}
\newcommand{\eg}{\emph{e.g.},}

\newcommand{\etal}{\emph{et al.}}
\newcommand{\Sec}{\S}

\DeclareAcronym{CAD}{
short=CAD,
long=computer-aided diagnosis
}

\DeclareAcronym{PXR}{
short=PXR,
long=pelvic X-ray
}
\DeclareAcronym{CXR}{
short=CXR,
long=chest X-ray
}
\DeclareAcronym{CNN}{
short=CNN,
long=convolutional neural network
}
\DeclareAcronym{AUC}{
short=AUCROC,
long=area under the ROC curve,
long-plural-form=areas under the ROC curve
}
\DeclareAcronym{ROC}{
short=ROC,
long=receiver operating characteristic 
}
\DeclareAcronym{ROI}{
short=ROI,
long=region of interest
}
\DeclareAcronym{FCN}{
short=FCN,
long=fully-convolutional network
}
\DeclareAcronym{MIL}{
short=MIL,
long=multiple instance learning
}
\DeclareAcronym{PACS}{
short=PACS,
long=picture archiving and communication system
}
\DeclareAcronym{ER}{
short=ER,
long=emergency room
}
\DeclareAcronym{SR}{
short=S@R$95$,
long=specificity at recall rate of $95\%$
}
\DeclareAcronym{RS}{
short=R@S$95$,
long=recall at specificity rate of $95\%$
}
\DeclareAcronym{PR}{
short=P@R$95$,
long=precision at recall rate of $95\%$
}
\DeclareAcronym{LSE}{
short=LSE,
long=log of the sum of the exponentials
}
\DeclareAcronym{BCE}{
short=BCE,
long=binary cross entropy
}
\DeclareAcronym{GAP}{
short=GAP,
long=global average pooling
}

\title{Weakly Supervised Universal Fracture Detection in Pelvic X-rays}
\author{\small Yirui Wang\textsuperscript{1}, Le Lu\textsuperscript{1}, Chi-Tung Cheng\textsuperscript{2}, Dakai Jin\textsuperscript{1}, Adam P. Harrison\textsuperscript{1}, \\ Jing Xiao\textsuperscript{3}, Chien-Hung Liao\textsuperscript{2}, Shun Miao\textsuperscript{1}}
\authorrunning{Y. Wang et al.}
\institute{\small \textsuperscript{1}PAII Inc., Bethesda, MD, USA \\ \textsuperscript{2}Chang Gung Memorial Hospital, Linkou, Taiwan, ROC \\ \textsuperscript{3}Ping An Technology, Shenzhen, China}

\begin{document}

\maketitle

\begin{abstract}
Hip and pelvic fractures are serious injuries with life-threatening complications. However, diagnostic errors of fractures in \acp{PXR} are very common, driving the demand for \ac{CAD} solutions. A major challenge lies in the fact that fractures are localized patterns that require localized analyses. Unfortunately, the \acp{PXR} residing in hospital \acl{PACS} do not typically specify \aclp{ROI}. In this paper, we propose a two-stage hip and pelvic fracture detection method that executes localized fracture classification using weakly supervised \ac{ROI} mining. The first stage uses a large capacity \acl{FCN}, \ie{} deep with high levels of abstraction, in a \acl{MIL} setting to automatically mine probable true positive and definite hard negative \acp{ROI} from the whole \ac{PXR} in the training data. The second stage trains a smaller capacity model, \ie{} shallower and more generalizable, with the mined \acp{ROI} to perform localized analyses to classify fractures. During inference, our method detects hip and pelvic fractures in one pass by chaining the probability outputs of the two stages together. We evaluate our method on $4\,410$ \acp{PXR}, reporting an \acl{AUC} value of $0.975$, the highest among state-of-the-art fracture detection methods. Moreover, we show that our two-stage approach can perform comparably to human physicians (even outperforming emergency physicians and surgeons), in a preliminary reader study of $23$ readers. 
\end{abstract}


\acresetall
\section{Introduction}
\label{sec:intro}

Hip and pelvic fractures belong to a frequent trauma injury category worldwide~\cite{johnell2004estimate}. Frontal \acp{PXR} are the standard imaging tool for diagnosing pelvic and hip fractures in the \ac{ER}. However, anatomical complexities and perspective projection distortions contribute to a high rate of diagnostic errors~\cite{chellam2016missed} that may delay treatment and increase patient care cost, morbidity, and mortality~\cite{tarrant2014preventable}. As such, an effective \ac{PXR} \ac{CAD} approach for \emph{both} pelvic and hip fractures is of high clinical interest, with the aim of reducing diagnostic errors and improving patient outcomes.  

Image-level labels are the only supervisory signal typically available in \ac{PACS} data. Thus, a widely adopted formulation for X-ray abnormality detection is a single-stage global classifier~\cite{badgeley2018deep, wang2017chestx8, rajpurkar2017chexnet, chen2018deep}. However, for \acp{PXR} this approach is challenged by the localized nature of fractures and the complexity of the surrounding anatomical regions. Moreover, such global classifiers can be prone to overfitting, as it is unlikely that a training dataset could capture the combinatorial complexity of configurations of fracture locations, orientations, and background contexts within the whole \ac{PXR}---this complexity is analogous to similar challenges within computer vision~\cite{DBLP:journals/corr/abs-1805-04025}. Indeed, for \emph{hip fractures alone}, Jim{\'e}nez-S{\'a}nchez \etal{} show that using localized \acp{ROI} produces significantly better F1 scores over a global approach~\cite{jimenez2018weakly} and Gale \etal{} achieve impressive \acp{AUC} of $0.994$ by first automatically extracting \acp{ROI} centered on the femoral neck~\cite{gale2017detecting}. These recent results bolster the argument for concentrating on local fracture patterns.

\begin{figure}[ht]
\centering
\small
\includegraphics[width=0.3\linewidth]{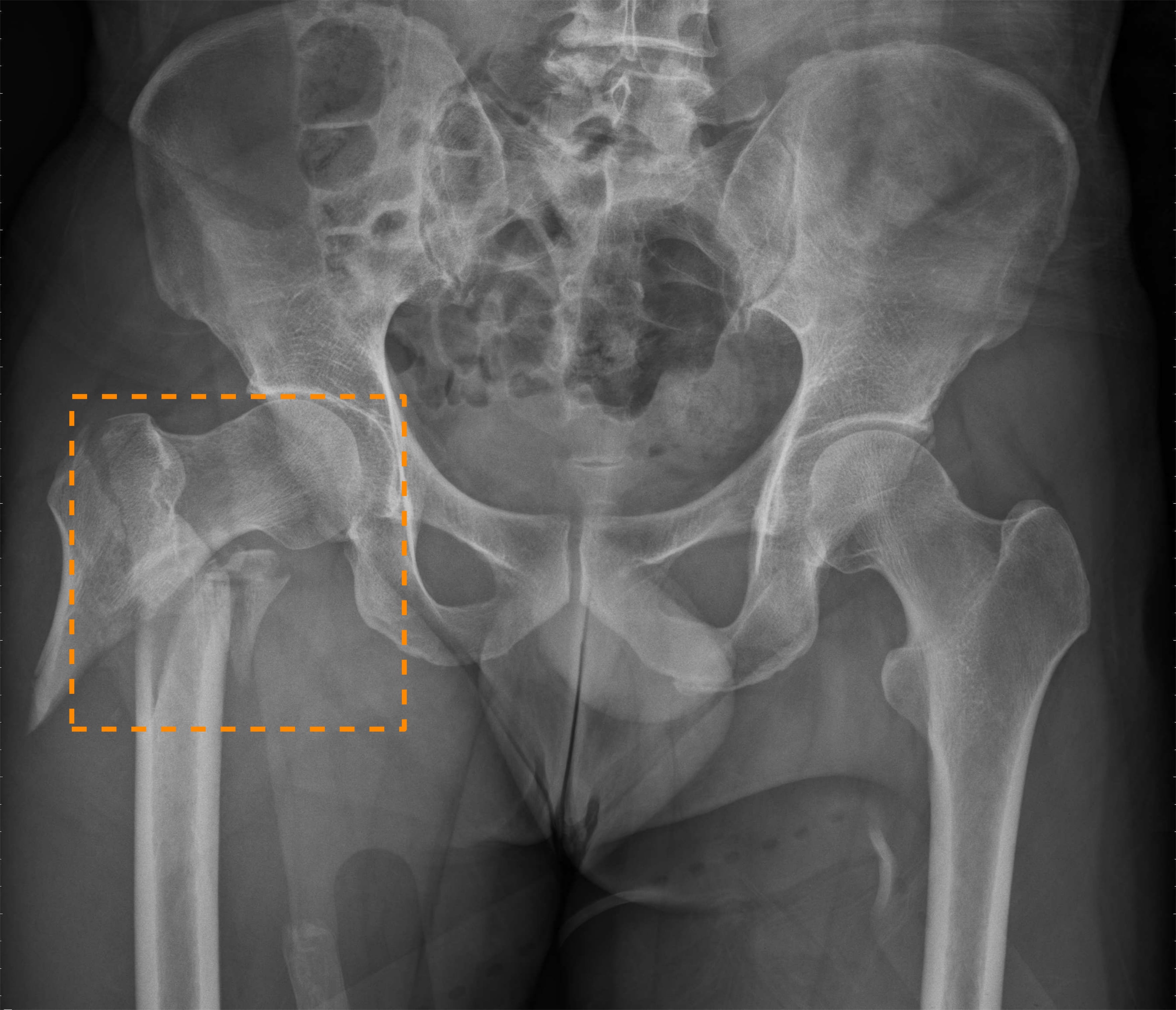}
\includegraphics[width=0.3\linewidth]{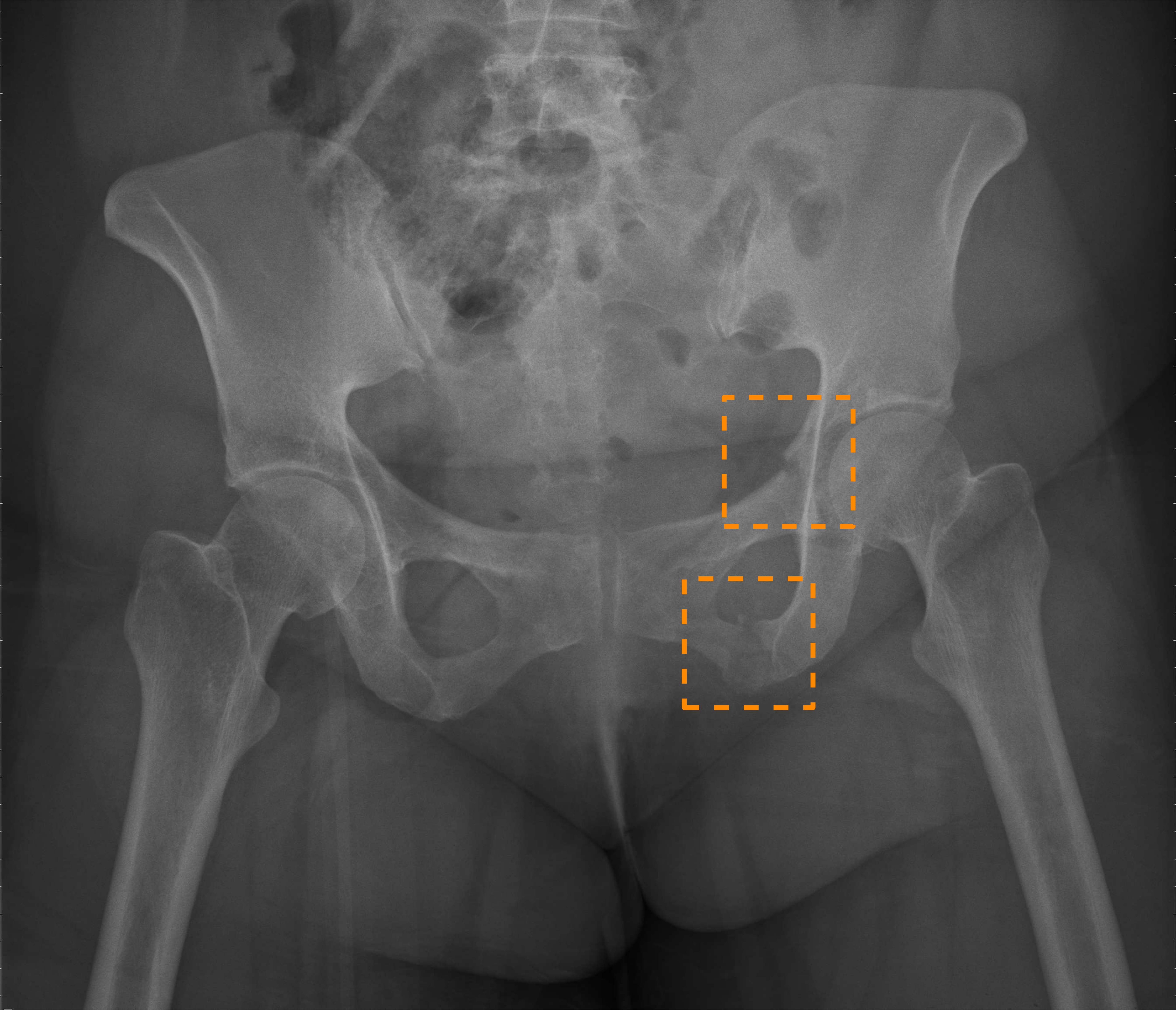}
\includegraphics[width=0.3\linewidth]{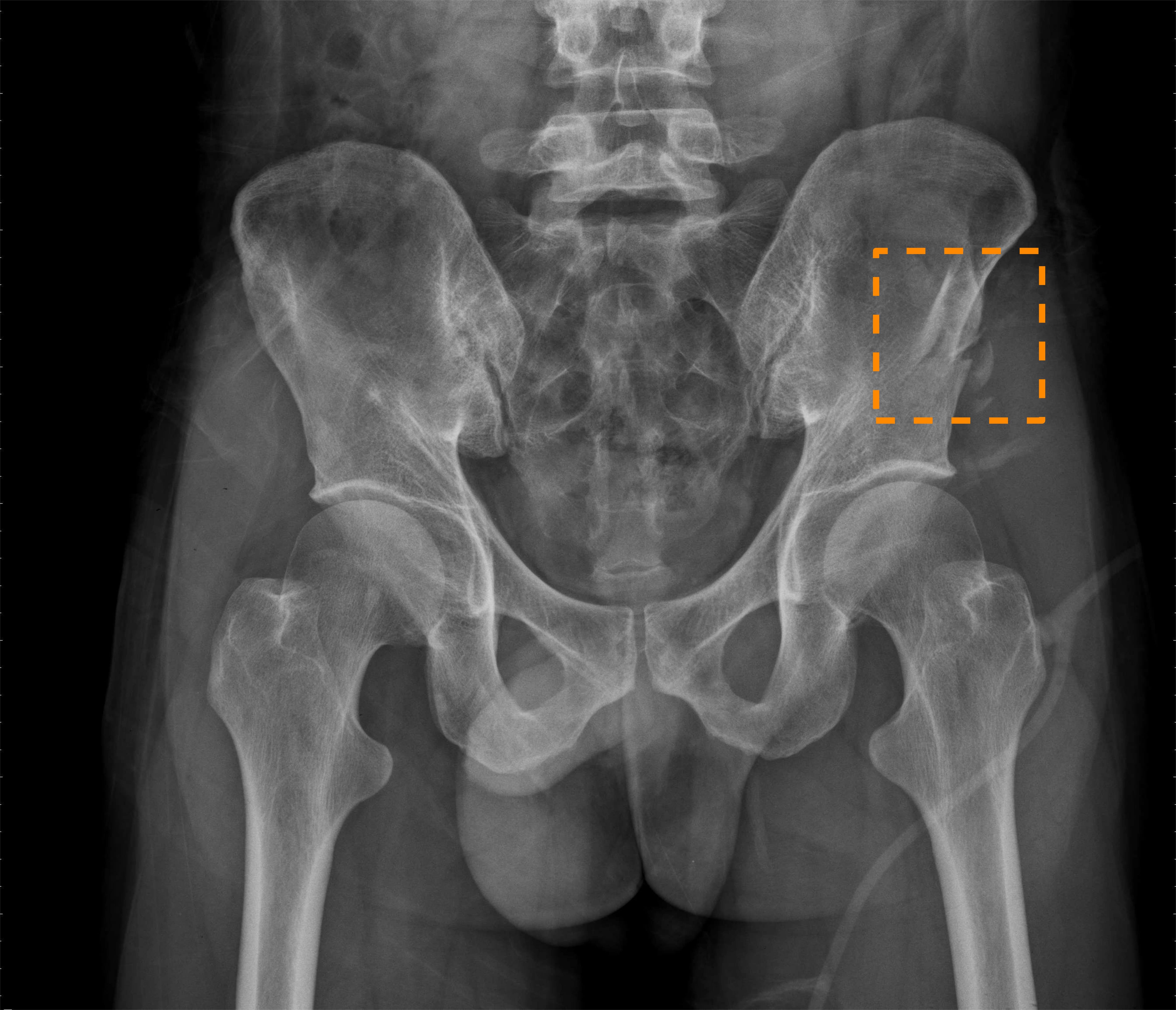}
\caption{Example \ac{PXR} images of hip and pelvic fractures. \textbf{(Left)} Hip fracture. \textbf{(Middle)} Superior and inferior pubic ramus fracture. \textbf{(Right)} Iliac wing fracture. }
\label{fig:fractures}
\end{figure}

Nonetheless, the above prior work all only focuses on diagnosing hip fractures and does not attempt to classify the more complex pelvic fractures (fractures in three pelvic bones: the ilium, ischium, and pubis). As Fig.~\ref{fig:fractures} illustrates, the makeup of pelvis fractures is much more complex, as there are a large variety of possible types with very different visual patterns at various locations. In addition, pelvic bones overlap with the lower abdomen, further confounding image patterns. Finally, unlike hip fractures, which occur at the femoral neck/head, pelvic fractures can occur anywhere on the large pelvis, both increasing the aforementioned image pattern combinatorial complexity and precluding automatic \ac{ROI} extraction based on anatomy alone, such as was done in prior work~\cite{gale2017detecting}. Thus, while using \ac{ROI}-based classification is even more desirable for pelvic fractures, it is paradoxically more challenging to extract said \acp{ROI}.

To bridge this gap, we propose a two-stage weakly supervised \ac{ROI} mining and subsequent classification method for \ac{PXR} fracture classification. In the first stage, we train a weakly-supervised, but high capacity, \ac{MIL} \ac{FCN} to mine local probable positive and definite hard negative \acp{ROI}. In the second stage, we use the mined \acp{ROI} to train a lower capacity network in a fully-supervised setting. During inference, the two networks are chained together to provide a complete classification solution. Experiments use a dataset of $4\,410$ \acp{PXR}, with only image-level labels, that we collected from the \ac{PACS} of Chang Gung Memorial Hospital. We show that single-stage classifiers, whether low- or high-capacity, are unable to match our two-stage approach. Our chained two-stage method outperforms the best single-stage alternative, with a \ac{SR} of $87.6\%$ compared to $80.9\%$, and corresponding improvements in \acp{AUC}. Moreover, using an independent reader study of $150$ patients, our system achieves an accuracy of $0.907$, which is equivalent to $23$ physicians. As such, we are the first to tackle automatic \ac{PXR} pelvis fracture classification and also the first to demonstrate diagnostic performance equivalent to human physicians for \emph{both} hip and pelvic fractures.

\section{Method}
\label{sec:method}



\begin{figure}[t!]
\centering
\includegraphics[width=\columnwidth]{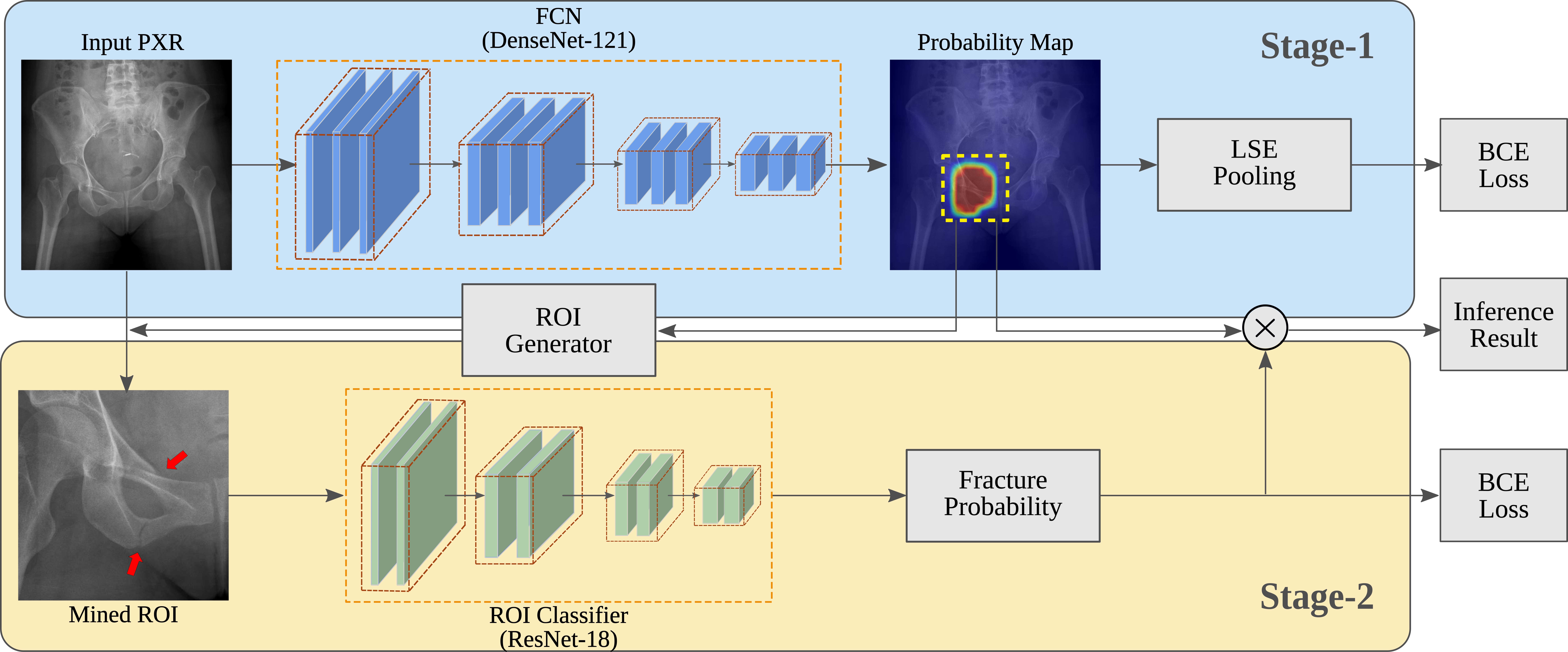}
\caption{The proposed two-stage fracture detection system. The first stage uses a large capacity \acs{MIL} \acs{FCN} model to perform fracture classification with weakly supervised \ac{ROI} localization. The second stage uses a smaller capacity model trained with the mined \acp{ROI} to perform localized classification. During inference, the two stages are chained together, with the second model applied on the \acp{ROI} proposed by the first model, to produce the final estimation.}
\label{fig:overview}
\end{figure}

Fig.~\ref{fig:overview} depicts the overall workflow of our chained two-stage pelvic and hip fracture detection method. We elaborate on the two stages below. 

\subsection{Weakly-Supervised \acs{ROI} Mining}
\label{subsec:stage1}

In the first stage, we train an \ac{FCN} using a deep \ac{MIL} formulation~\cite{yao2018weakly}, employing the large-capacity DenseNet-121~\cite{huang2017densely} network as backbone. The DenseNet-121 features are then processed using a $1\times 1$ convolutional layer and a sigmoid activation to produce a probability map. Owing to the localized properties of \acp{FCN}, each value of the probability map can be interpreted as the probability of fracture in the corresponding region in the input \ac{PXR}. The maximum value would then represent the probability of fracture within the entire image. Instead, we use \ac{LSE} pooling, which is a differentiable approximation of max pooling, given by  
\begin{equation}
   LSE(S)=\frac{1}{r} \cdot {\rm log}\left[ \frac{1}{\vert S \vert } \cdot \sum_{(i,j) \in \boldsymbol{S}} {\rm exp} \left( r \cdot p_{ij} \right) \right]\mathrm{,}
\end{equation}
where $\{p_{ij}\}$ is the probability map, and $r$ is a hyper-parameter controlling the behavior of \ac{LSE} between max pooling ($r \to \infty$) and average pooling ($r \to 0$). With the pooled global probability, \ac{BCE} loss is calculated against the image level label, and is used to train the network. While, this formulation has been applied directly for weakly supervised abnormality detection in \acp{CXR}~\cite{yao2018weakly}, as we show in our results this approach's performance is limited for hip and pelvic fracture detection. Therefore, we use the \ac{FCN} as a proposal generator to mine \acp{ROI} from the training data. 

To mine \acp{ROI} from the training data to train a localized classification model, we first create an image-level classifier using $p' = \max_{i,j \in S} p_{ij}$,
and select a threshold $\hat{p}$ corresponding to a high sensitivity on the training data (we use $99\%$ in our experiments). We then extract up to $K=5$ \acp{ROI} from each \ac{PXR} in the training data in every training epoch of the second stage model. Specifically, for \acp{PXR} with positive ground-truth image-level labels, \ie{} with fracture(s), up to $K$ locations are randomly selected from
\begin{equation}
    S' = \{i,j \vert p_{ij} > \hat{p}\}.
\end{equation}
These \acp{ROI} are labeled as probable fracture positive. For \acp{PXR} with negative ground-truth image-level labels, \ie{} no fractures, the same \ac{ROI} extraction strategy selects up to $K$ \acp{ROI}. These are considered as definite hard negatives. If there are less than $K$ hard negatives extracted, additional negative \acp{ROI} are randomly extracted from the \ac{PXR} to make up the total. The \acp{ROI} produced using the above strategy contains probable positives, hard negatives, and easy negatives. Although this approach adds a degree of label noise due to the probable positive \acs{ROI}, as we outline in the following, this comes with the added benefits of using a subsequent localized and more generalizable \ac{ROI} classifier.

\subsection{Fracture \acs{ROI} Classification}
\label{subsec:stage2}

In the second stage, we use the \acp{ROI} mined from the first stage as training data for a fully supervised localized classification network. Since the positive samples are mostly \acp{ROI} around fractures with limited background context, the visual patterns of fractures become more dominant, simplifying the classification task. In addition, the distribution of mined \acp{ROI} are heavily weighted toward hard negatives, \ie{} false positive regions from the first stage. This concentrates the modeling power of the second-stage classifier on differentiating these difficult/confusing fracture-like patterns. As a result, we are able to train a smaller capacity network, \eg{} ResNet-18~\cite{he2016deep}, to reliably classify the \acp{ROI}, which is more generalizable and less prone to overfitting compared to a high-capacity network modeling the entire \ac{PXR}.

During inference, the two stages are chained together to provide a complete solution. The first stage \ac{FCN} acts as a proposal generator, and the highest value from the probability map $\{p_{i,j}\}$ is selected, denoted as $p_{s1}$, along with the corresponding \ac{ROI}. The second stage classifier is then applied on the proposed \ac{ROI} to produce a fracture probability score, denoted $p_{s2}$. The final image-level probability of fracture is computed by multiplying the two probability scores, $p_{s1} \cdot p_{s2}$. As such, we use the second stage classifier as a filter to reject false positives from the first stage.

{\bf Hip/pelvic fracture differentiation:} Our two-stage method detects hip and pelvic fractures as one class, because the most important goal of \ac{PXR} \ac{CAD} is to detect fractures. Using one universal fracture class also helps to prevent the model from picking up co-occurrence relationships between hip/pelvic fractures that may be overly represented from the current training data~\cite{DBLP:journals/corr/abs-1805-04025}. In scenarios where hip and pelvic fractures do need to be differentiated, \eg{} automatic medical image reporting, an additional classification output node can be added to the second stage model. Similar to the hierarchical classification schemes~\cite{chen2018deep}, the new node is trained only on positive fracture \acp{ROI} mined in the first stage. Like fracture classification, during inference hip/pelvic fracture differentiation can be obtained in one feed-forward pass of the network.

\section{Experiments and Results}
\label{sec:experiments}

\begin{figure}[t]
    \centering
    \begin{subfigure}[b]{0.47\textwidth}
        \includegraphics[width=\textwidth]{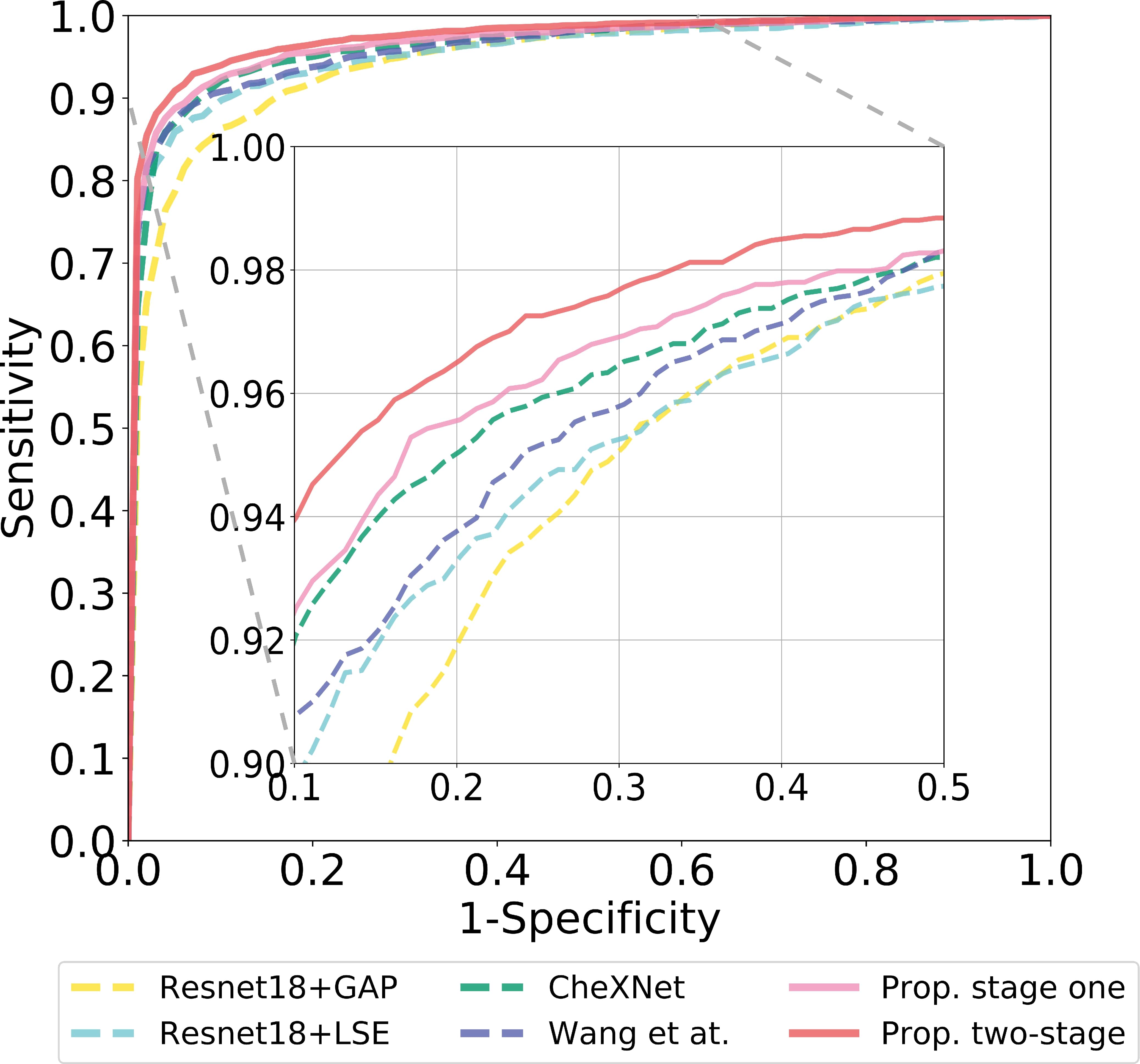}
    \end{subfigure}
    \begin{subfigure}[b]{0.47\textwidth}
        \includegraphics[width=\textwidth]{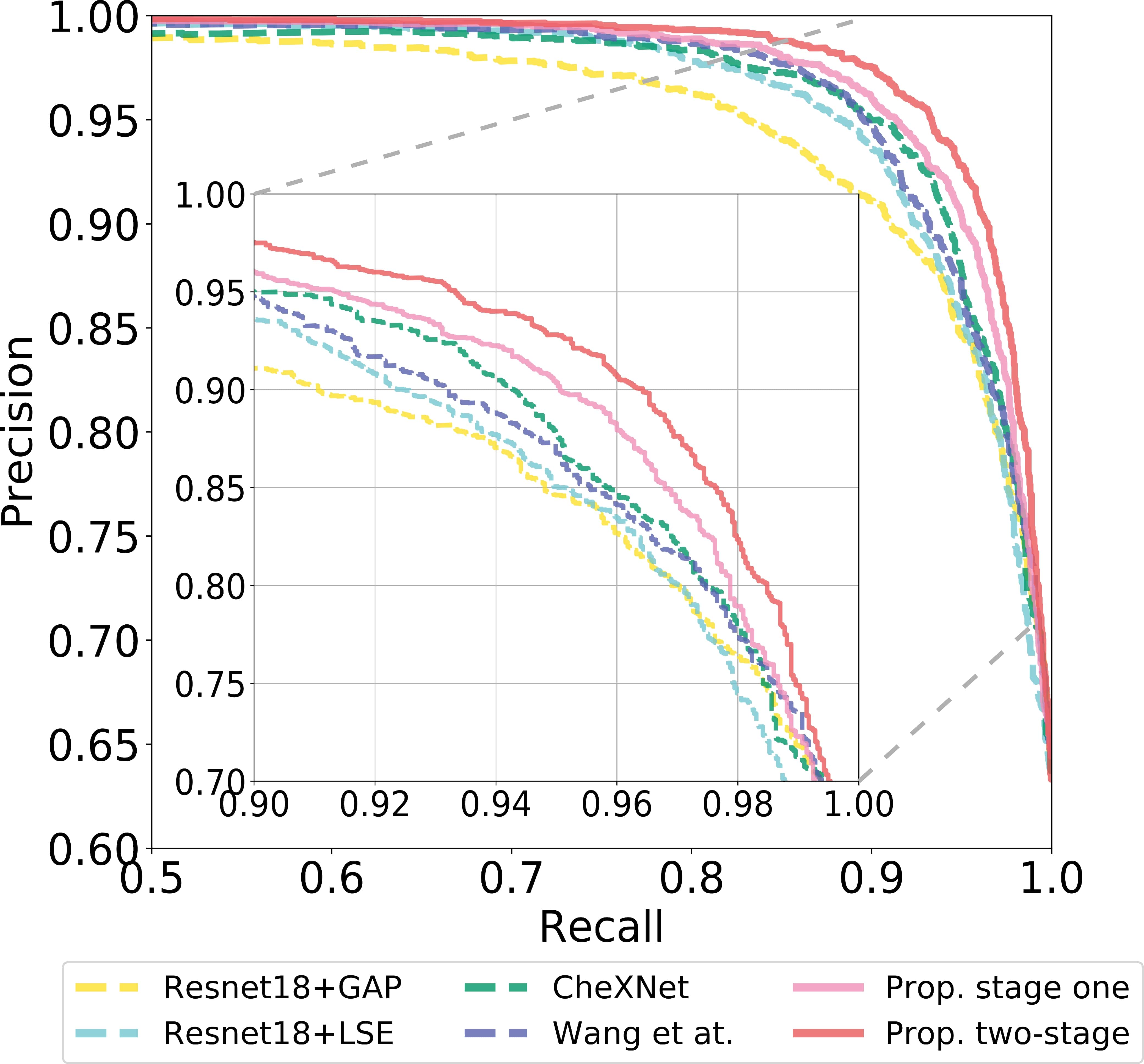}
    \end{subfigure}
    \caption{Comparison of ROC \textbf{(left)} and Precision-Recall curve \textbf{(right)}}
    \label{fig:ROC}
\end{figure}

We evaluate our framework using \ac{PXR} images collected from the \ac{PACS} of Chang Gung Memorial Hospital, corresponding to patients in the trauma registry. We resized all images to $961\times961$ pixels. The final dataset consisted of $4\,410$ images, including $2\,776$ images with fractures ($1\,975$ and $801$ hip and pelvic fractures, respectively). Besides this dataset, we also collected an independent \ac{PXR} dataset, containing $150$ cases ($50$ hip fractures, $50$ pelvic fractures, and $50$ no findings) for a reader study comparing our approach with that of $23$ physicians. 

We use ImageNet pre-trained weights to initialize the networks in both stages. The Adam
optimizer was used to train both models for $100$ epochs with a batch size of $8$ and a starting learning rate of $10^{-5}$ reduced by a factor of $10$ upon plateaus. In addition to \ac{AUC}, we measure \acf{SR}, \ac{PR} and \ac{RS}, which help highlight differences in performance under demanding expectations for recall/sensitivity and specificity, respectively. 

\subsection{Comparison to Prior Work}
We evaluate \emph{general fracture} classification performance using five-fold cross-validation with a $70\%/10\%/20\%$ training, validation, and testing split, respectively. We compare against the single-stage high-capacity approaches of CheXNet~\cite{rajpurkar2017chexnet} and Wang \etal{}~\cite{wang2017chestx8}, both of which use DenseNet-121 as backbones and apply \ac{GAP} and \ac{LSE} pooling, respectively. Note, that unlike our first stage of \Sec\ref{subsec:stage1}, the pooling is applied to the last feature map. We also compare against the single-stage lower-capacity model of ResNet-18, using both \ac{GAP} and \ac{LSE} pooling heads. 

\begin{table}[t]
	\small
	\caption{Five-fold cross validation of fracture classification on $4\,410$ \acp{PXR}.}
	\label{tab:overallRank}
	\centering
	\begin{threeparttable}[b]
		\begin{tabular*}{\textwidth}{l @{\extracolsep{\fill}} cccc}
			\toprule
			Method & AUC & \ac{SR} & \ac{RS} & \ac{PR}\\ 
			\midrule
			ResNet18-GAP & 0.946 & 0.706 & 0.786 & 0.846\\
			ResNet18-LSE & 0.956 & 0.723 & 0.859 & 0.851\\
			CheXNet~\cite{rajpurkar2017chexnet} & 0.962 & 0.809 & 0.870 & 0.876\\
			Wang \etal{}~\cite{wang2017chestx8} & 0.962 & 0.752 & 0.875 & 0.867\\
			\midrule
			Prop. single-stage & 0.968 & 0.825 & 0.888 & 0.903\\
			Prop. two-stage & \textbf{0.975} & \textbf{0.876} & \textbf{0.909} & \textbf{0.928}\\
			\bottomrule
		\end{tabular*}
	\end{threeparttable} 
\end{table}

Fig.~\ref{fig:ROC} and Tbl.~\ref{tab:overallRank} quantitatively summarizes these experiments. As can be seen, all lower-capacity models fare relatively poorly, demonstrating the need to use more descriptive models for global \ac{PXR} interpretation. On the other hand, the first stage of our proposed method achieves an \ac{AUC} of $0.968$, compared to the $0.962$ achieved by the state-of-the-art single-stage methods~\cite{rajpurkar2017chexnet,wang2017chestx8}, demonstrating that our single-stage approach using deep \ac{MIL} can already outperform prior art. 
With the second stage, our method was able to further improve the \ac{AUC} to $0.975$, the highest among all evaluated methods. This corresponds to improvements of $12.4\%$ ($3.4\%$), $6.7\%$ ($3.9\%$), and $5.1\%$ ($2.1\%$) in \ac{SR} (\ac{RS}) over Wang \etal{}~\cite{wang2017chestx8}, CheXNet~\cite{rajpurkar2017chexnet}, and our single-stage model respectively. These are highly impactful boosts in  performance, demonstrating that under high demands of recall and specificity our chained approach can provide drastic improvements. 

At our two-stage method achieves a \ac{PR} of $0.928$, measuring $9.7\%$, $9.0\%$, $5.9\%$ and $7.0\%$ improvements over the two low-capacity baseline models, ResNet18-GAP and ResNet18-LSE, and two high-capacity baseline models CheXNet~\cite{rajpurkar2017chexnet} and Wang \etal{}~\cite{wang2017chestx8}, respectively. Please note that the actual prevalence of fractures in clinical environments can be lower than our data, which will result in lower precisions for all methods. Nonetheless, the performance ranking would likely to remain, and the improvements of our method over the baselines are expected to be even more significant with a lower prevalence.

In addition, we evaluate the label accuracy of mined probable positive \acp{ROI} on $438$ \acp{PXR} with fracture location annotations, and report a high accuracy of $0.925$, demonstrating the effectiveness of the proposed weakly-supervised \ac{ROI} mining scheme. We also evaluate the hip/pelvic fracture classification performance, and report a high average accuracy of $0.980$ over the five-fold cross validation. 

\subsection{Reader Study}
We conduct reader study to compare performance on $150$ \acp{PXR} with $23$ human physicians recruited from the surgical ($11$), orthopedics ($4$), \ac{ER} ($6$) and radiology ($2$) departments. For every \ac{PXR}, physicians were asked to choose from three options: hip fracture, pelvic fracture or no finding. To provide a fair comparison, we used the add-on fracture-type classification output node described in \Sec\ref{subsec:stage2}, which can differentiate between hip and pelvic fractures, matching the three-class classification performed by the readers.

\begin{table}[t]
	\small
	\caption{Algorithm and physician performances in a clinical study on 150 \acp{PXR}.}
	\label{tab:human}
	\centering
	\begin{tabular*}{\textwidth}{l @{\extracolsep{\fill}} ccccc}
		\toprule
		& \multirow{2}{*}{Accuracy} & \multicolumn{2}{c}{Hip Fracture} & \multicolumn{2}{c}{Pelvic Fracture}\\ \cmidrule{3-4} \cmidrule{5-6}
		& & Sensitivity & Specificity & Sensitivity & Specificity \\
		\midrule
		\acs{ER} physician & 0.881 & 0.983 & 0.937 & 0.813 & 0.955 \\
		Surgeon & 0.855 & 0.931 & 0.928 & 0.829 & 0.932 \\
		Orthopedics specialist & 0.932 & 1.000 & 0.953 & 0.905 & 0.990 \\
		Radiologist & 0.930 & 0.990 & 0.965 & 0.870 & 0.995 \\ \midrule
		Physician average & 0.882 & 0.962 & 0.938 & 0.842 & 0.953 \\
		Our method & 0.907 & 0.960 & 0.980 & 0.840  & 0.960 \\
		\bottomrule
	\end{tabular*} 
\end{table}

Tbl.~\ref{tab:human} quantitatively summarizes the reader study results. As shown in the results, our method performs comparably to the average physician performance on this dataset, reporting an accuracy of $0.907$ compared to $0.882$, with higher levels of specificity. Examining the physician specialities in isolation, our method outperforms \ac{ER} physicians and surgeons, while not performing as well as the orthopedic specialists and radiologists. Of note, is that when trauma patients are sent to the \ac{ER}, it is common that only \ac{ER} physicians or surgeons are available to make immediate diagnostic and treatment decisions. As such, the reader study suggests that our approach may be an effective aid for \ac{PXR} fracture diagnosis in the high-stress \ac{ER} environment. 

\section{Conclusion}
\label{sec:conclusion}

We introduced a chained two-stage method for universal fracture detection in \acp{PXR}, consisting of a weakly supervised fracture \ac{ROI} mining stage and a localized fracture \ac{ROI} classification stage. Experiments show that our method can significantly outperform prior works via five-fold cross validation on $4\,410$ \acp{PXR}. Moreover, a preliminary reader study on $150$ \acp{PXR} involving $23$ physicians suggests that our method can perform equivalently to human physicians. Thus, our approach represents an important step forward in automated pelvic and hip fracture diagnosis for \ac{ER} environments.

\bibliographystyle{splncs04}
\small{\bibliography{refs}}
\end{document}